\documentclass[times,twocolumn,final]{elsarticle}

\usepackage[table]{xcolor} 
\usepackage{medima}
\usepackage{framed,multirow}

\usepackage{amssymb}
\usepackage{latexsym}

\usepackage{booktabs}
\usepackage{amsmath}
\usepackage[linesnumbered,ruled]{algorithm2e}

\usepackage{url}
\usepackage{xcolor}
\usepackage{colortbl}
\usepackage{makecell}

\usepackage{hyperref}

\definecolor{newcolor}{rgb}{.8,.349,.1}
\definecolor{newred}{HTML}{EF3B69}
\definecolor{newgreen}{HTML}{42BB94}
\definecolor{newyellow}{HTML}{FFD160}
\definecolor{newblue}{HTML}{034B77}
\definecolor{lightyellow}{rgb}{1.0, 1.0, 0.88}

\journal{Medical Image Analysis}

\begin{document}

\verso{Wang \textit{et~al.}}

\begin{frontmatter}

\title{Task-oriented Embedding Counts: Heuristic Clustering-driven Feature Fine-tuning for Whole Slide Image Classification}

\author[1]{Xuenian Wang\fnref{fn1}}
\author[1]{Shanshan Shi\fnref{fn1}}
\author[1]{Renao Yan\fnref{fn1}\corref{cor1}}
\author[1]{Qiehe Sun}
\author[2]{Lianghui Zhu}
\author[1]{Tian Guan}
\author[1,3]{Yonghong He\corref{cor1}}
\cortext[cor1]{Corresponding authors. E-mail addresses: yra21@mails.tsinghua.edu.cn (R. Yan), heyh@sz.tsinghua.edu.cn (Y. He).}
\fntext[fn1]{Equal contribution.}
\address[1]{Shenzhen International Graduate School, Tsinghua Unversity, Beijing, China}
\address[2]{Shenzhen Shengqiang Technology Co., Ltd., Shenzhen, China}
\address[3]{Greater Bay Area National Center of Technology Innovation, Guangzhou, China}

\received{11 April 2024}

\begin{abstract}
In the field of whole slide image (WSI) classification, multiple instance learning (MIL) serves as a promising approach, commonly decoupled into feature extraction and aggregation. In this paradigm, our observation reveals that discriminative embeddings are crucial for aggregation to the final prediction. Among all feature updating strategies, task-oriented ones can capture characteristics specifically for certain tasks. However, they can be prone to overfitting and contaminated by samples assigned with noisy labels. To address this issue, we propose a heuristic clustering-driven feature fine-tuning method (HC-FT) to enhance the performance of multiple instance learning by providing purified positive and hard negative samples. Our method first employs a well-trained MIL model to evaluate the confidence of patches. Then, patches with high confidence are marked as positive samples, while the remaining patches are used to identify crucial negative samples. After two rounds of heuristic clustering and selection, purified positive and hard negative samples are obtained to facilitate feature fine-tuning. The proposed method is evaluated on both CAMELYON16 and BRACS datasets, achieving an AUC of 97.13\% and 85.85\%, respectively, consistently outperforming all compared methods.
\end{abstract}


\end{frontmatter}



\section{Introduction}

Recent advancements \citep{song_artificial_2023} in scanning systems, imaging technologies, and storage devices have enabled a substantial increase in the production of digital pathology slides, commonly known as whole slide images (WSI). This ever-increasing volume of WSI results in a manpower shortage dilemma in pathological analysis. To alleviate this dilemma, leveraging computerized algorithmic methods for analyzing vast quantities of WSI has emerged as a promising solution \citep{srinidhi_deep_2021, louis_computational_2016}. The process of using artificial intelligence to analyze WSI is commonly referred to as computational pathology \citep{abels2019computational}. Computational pathology can provide patients and clinicians with more objective diagnostic and prognostic results, in the field of tumor regions detection\citep{7727519}, immunohistochemistry scores \citep{article}, cancer staging \citep{article1, inproceedings}, virtual staining \citep{yan2023unpaired}, mitosis detection \citep{DBLP:journals/corr/ShahRSW16, article2}, and glandular segmentation \citep{DBLP:journals/corr/0011QYH16, article3,86de9d393ab141fabc0dd585e66647fe}.

One of the central aims of computational pathology is to discern disease types, a task commonly referred to as WSI classification. Diverging from natural image classification tasks, one WSI possesses gigapixels, thus directly feeding a WSI into a convolutional neural network (CNN) would result in memory overflow. One of the typical solutions involves segmenting WSI into patches to reduce the processing burden. However, this approach brings another challenge: given the necessity for specialized domain knowledge in pathological image analysis and the extensive number of patches, acquiring patch-level annotations is impractical.

To tackle this challenge, multiple instance learning (MIL) \citep{ABMIL,clam,DSMIL,DTFD,ACMIL,WSIFT,BCL,zhu2023accurate,yan2023shapley,tang2023multiple,qiehe2024nciemil,li2024dynamic,chu2024retmil,chen2024camil}, a type of weakly supervised learning, is extensively utilized for WSI classification due to its capability \citep{dietterich1997solving,9269335}, where the whole slide image is treated as a bag of numerous unlabeled instances (patches). Originally designed for binary classification, MIL stipulates that a bag is considered positive if at least one instance within it is positive; otherwise, it is negative. Multiple instance learning networks typically comprise three components: a feature encoder that embeds patches into low-dimensional feature vectors, a MIL aggregator that integrates patch-level features to bag-level features, and a multi-layer perception (MLP) that predicts the category of the bag-level features. In natural image classification, MIL training is usually trained in an end-to-end paradigm. However, directly applying this paradigm to pathological images poses challenges \citep{hou2016patchbased,sharma2021clustertoconquer,ABMIL,clam} resulting from the inability to store all intermediate feature maps in parallel reported in \citep{BCL}.
\begin{figure}[t!]
    \centering
    \includegraphics[width=0.95\linewidth]{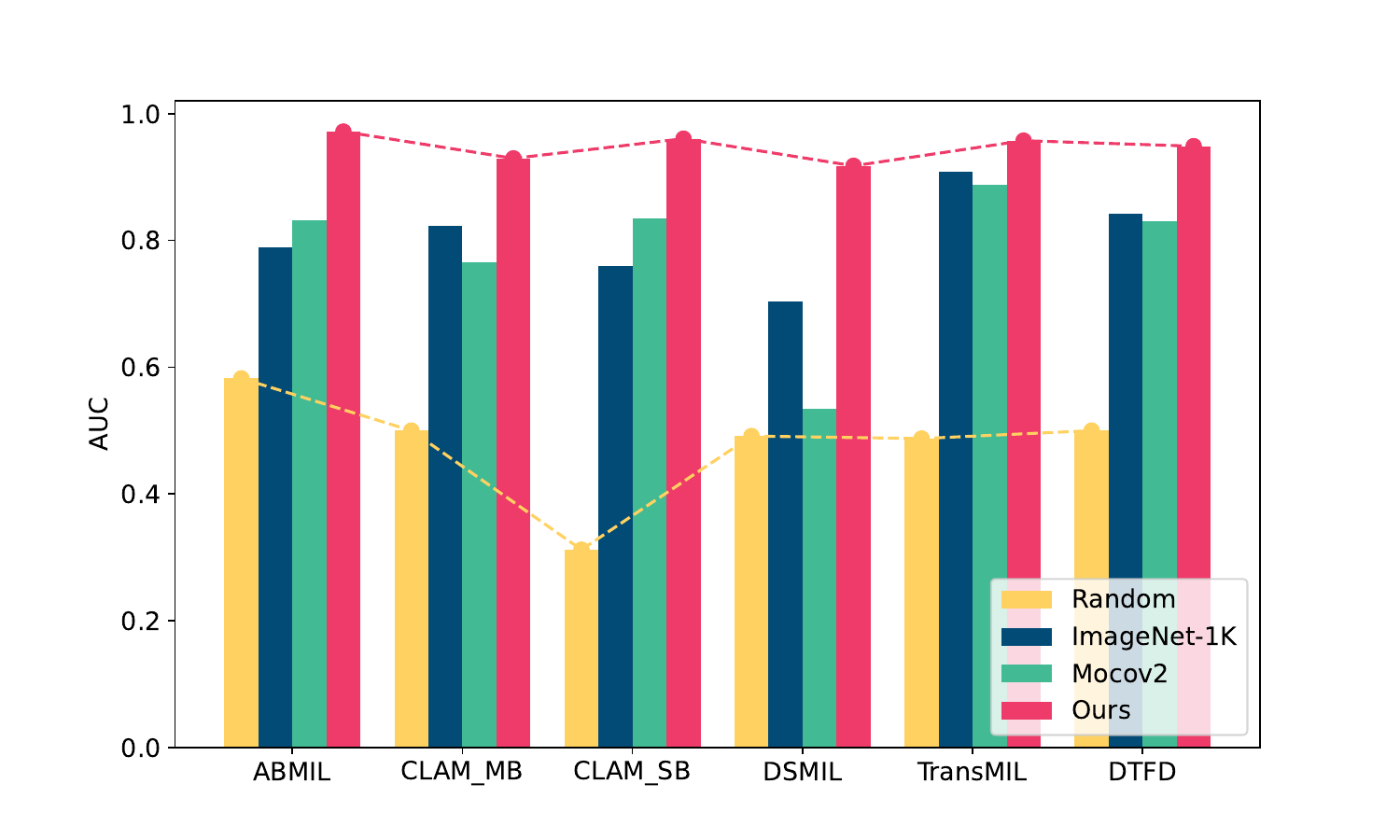}
    \caption{AUC performance of various MIL models based on different pre-trained weights on the CAMELYON16 test dataset.}
    \label{fig1}
\end{figure}

A common approach involves training the feature encoder and the aggregator separately. This typically entails using a pre-trained encoder to extract features, which is then held constant while solely optimizing the MIL aggregator. Currently, several strategies exist for updating the feature encoder, including initializing weights randomly, employing weights pre-trained on natural images through supervised learning, utilizing weights pre-trained on homogeneous images via self-supervised learning (SSL), and leveraging weights pre-trained on task-oriented images via supervised learning. Our observation on CAMELYON16, as depicted in Fig. \ref{fig1}, demonstrates that employing randomly initialized weights for the feature encoder yields unsatisfactory results. Conversely, utilizing pre-trained weights from ImageNet-1k tends to be a more effective choice. However, due to the inherent differences between natural and pathological images, these pre-trained weights may not be optimal for feature extraction in pathological contexts \citep{WSIFT}. Alternatively, utilizing weights pre-trained via self-supervised learning helps bridge the domain gap, resulting in superior performance in ABMIL and CLAM\_SB. Nonetheless, this approach may yield suboptimal results as it lacks task-specific information. Furthermore, these non-task-oriented methods are significantly influenced by the efficacy of the aggregator. 

In response to these challenges, task-oriented feature fine-tuning has emerged as a profitable approach, which facilitates the optimization of the encoder using task-specific information. In this paradigm, the attention score introduced by \citep{ABMIL} appears as a logical choice to distill patch-level information from bag-level predictions. However, \citep{yan2023shapley} suggests that extreme attention distribution and imprecise focus can lead to model overfitting, particularly when updating the feature encoder based on samples with noisy supervision. To mitigate these issues, some methods identify instances with the lowest attention scores as negative samples for further training. While this method is limited by the inability to identify crucial hard negative samples.

To address the aforementioned challenges, we propose HC-FT which contains a heuristic clustering strategy that merges conventional clustering processes with the pseudo label strategy for task-oriented feature fine-tuning. This strategy allows us to clean positive samples and detect hard negative samples. The cleaning process tackles the issue of potential encoder overfitting to noisy data observed in previous methods. Simultaneously, the detection of hard negative samples enhances feature discrimination and mitigates the problem of overly concentrated attention. In conclusion, our contribution is as follows:

\begin{itemize}
\item In the realm of multiple instance learning for WSI classification, our observation indicates the significance of the encoder. We found that discriminative feature embeddings can yield effective results across various aggregators.
\item To address potential performance degradation caused by noisy samples, we propose a heuristic clustering strategy aimed at ensuring the purity of information to update the encoder. Additionally, we incorporate hard negative samples to enhance the feature discrimination. 
\item Extensive experiments conducted on two public datasets demonstrate the superiority of our method. Furthermore, visualization results provide compelling evidence that our features possess considerable discrimination.
\end{itemize}

\section{Related works}
\subsection{Multiple instance learning for WSI classification}
In general, MIL approaches are categorized into instance-based and embedding-based methods. Instance-based methods focus on acquiring labels for individual instances and aggregating them for bag-level classification. Bag-level prediction is aggregated by the probability of all patches with Mean or Max-pooling \citep{campanella2019clinical}. As for embedding-based methods, they embed all patch-level features to form the feature set of bags, directly predicting the bag's features. Due to their superior performance and enhanced interpretability, these methods are widely used in WSI classification. 

In the realm of embedding-based MIL, a notable advancement is the attention-based MIL (ABMIL) introduced by \citep{ABMIL}. This pioneering method leverages gated attention mechanisms to assign learnable weights to features at the patch level, marking a significant milestone in the evolution of MIL techniques. Concurrently, \citep{clam} have enhanced performance by employing attention mechanisms for the selective sampling of positive and negative instances, coupled with cluster constraints to mine patch-level information and integrate bag-level predictions. Furthermore, \citep{DSMIL} identifies key instances and their relations with other instances for bag-level embedding, this approach also underscores the potential of self-supervised methods in enhancing model performance. \citep{ACMIL} replaced the self-attention mechanism with a multi-head attention mechanism and a masking strategy is applied to patches with high attention to mitigate issues of excessive attention concentration and model overfitting.

However, several MIL models consider only the semantic information and neglect the spatial relationships between patches. This results in the loss of valuable positional information and overlooks the significance of context in pathological diagnosis. This oversight has been addressed by \citep{DSMIL, TransMIL, IBMIL}. \citep{DSMIL} have utilized multi-scale information to construct pyramid-structured data, thereby implicitly incorporating spatial information. Expanding upon this, \citep{TransMIL} have proposed a model based on the transformer architecture to map the interrelations between patches, while \citep{IBMIL} have employed backdoor adjustment for intervention training to mitigate biases introduced by contextual priors. Another method \citep{graph0, Zhao_2020_CVPR, graph2, li2024dynamic} to represent the spatial relationships between different patches is through graph theory, where \citep{Zhao_2020_CVPR} have integrated MIL with deep graph convolutional networks, demonstrating superior performance in predicting lymph node metastasis. Similarly, \citep{graph2} have applied graph neural networks to learn the relationships between patches, using graph pooling to infer patches of higher relevance automatically.

Other researches concentrate on the representation of bags, \citep{bergner2023iterative} and \citep{kong2021efficient} identifying prominent patches to form bag embeddings. \citep{DTFD} introduced the concept of pseudo bags, developing a two-layer distillation framework that forms diverse bags through distinct distillation methods.

These approaches have to decouple the training of the feature encoder and the MIL aggregator to avoid the memory bottleneck \citep{BCL}, resulting in performance degradation.

\subsection{Self-supervised learning}

Self-supervised learning, an emergent training paradigm within representation learning \citep{BT,MoCo,mocov2,MAE,mocov3,DINO,swav}. Unlike supervised learning, which is constrained by the availability of labeled data, self-supervised learning can learn from a vast amount of unlabeled data \citep{chen2020simple,misra2019selfsupervised}. In the domain of computer vision, self-supervised learning is capable of generating universal visual features that can even surpass the performance of models trained on labeled data, even on highly competitive benchmarks like ImageNet-1K \citep{tomasev2022pushing,MoCo}. It is predominantly employed in computational pathology to construct patch embeddings. Given its capacity to derive domain-specific patch embeddings without the need for annotated data, leveraging extensive pathological data for SSL to obtain pre-trained weights emerges as an effective strategy.

The proliferation of patch-level self-supervised training paradigms within computational pathology has been noteworthy \citep{DSMIL,ciga2021self,azizi2022robust}. For instance, contrastive learning facilitates the mapping of patches and their augmented versions 
\citep{MoCo} (typically through pathology-relevant transformations such as random rotation, cropping, and stain jitter) to analogous embeddings. By utilizing pretext tasks, such as discerning correspondences between global and local-level details \citep{Chen_2022_CVPR} or reconstructing randomly masked areas within patches \citep{MAE}, as well as leveraging similarities between patches from the same whole slide image or those sharing labels \citep{ctrans}. Self-supervised training methods, which learn similar spatial semantic information, are increasingly used for pre-trained feature encoders. Despite their ability to extract valuable information without annotations, SSL approaches require massive data and computational resources. As a result, a common approach is to use publicly available pre-trained weights on pathological datasets. Yet, the effectiveness of this strategy is limited due to variations in feature spaces across different tasks.
\subsection{End-to-end training and fine-tuning}

Early research has explored end-to-end training of MIL methods to update features. One approach centered around optimizing the training workflow, \citep{chenend} leveraged unified memory architectures, allowing GPUs direct access to host memory for streamlined end-to-end model training. \citep{9380553} capitalized on the locality of convolution operations in CNN to diminish memory demands, \citep{princkend2} also introduced an enhanced streaming CNN technique, tailored to optimize whole slide images for end-to-end training. These strategies have not achieved widespread adoption due to their computationally demanding nature. An alternative perspective focuses directly on WSIs. \citep{sharma2021clustertoconquer} developed an end-to-end training framework utilizing patch clustering, and \citep{wsiyasuo} applied unsupervised techniques to compress WSIs. While, in theory, end-to-end training models are posited to deliver optimal outcomes, their propensity for overfitting hampers their generalizability.

Fine-tuning is indispensable for acquiring task-oriented features while it is unavailable for direct use in WSI classification due to memory bottlenecks \citep{BCL}. A pragmatic strategy involves updating the feature encoder with a subset of samples. This task-oriented method iteratively transfers supervised information from one round to the next to update feature representations. For example, \citep{wulczyn2020deep,wulczyn2021interpretable} sampled patches from WSIs during training. \citep{WENO} proposed a weakly supervised knowledge distillation concept to update the feature encoder using soft labels while discarding easily classified samples to construct challenging pseudo bags. \citep{BCL} amalgamated bayesian and collaborative learning principles, applying a quality-aware strategy to update the feature encoder with pseudo labels. \citep{cao2023e2efp} engaged in iterative sampling of patches from WSIs to train the feature pyramid encoder, and \citep{tang2023multiple} employed masking on patches with high attention scores, compelling teacher-student models to learn more discriminative features. Although these methods successfully updated the feature encoder, the information used to update the feature encoder may be noisy, and the mask method to implicitly search for hard negative samples may cause the model to lose some key information from high attention patches.

\section{Method}

\subsection{Problem statement}

In a dataset of whole slide images with bag-level labels, denoted as $\mathcal{D}=\{(X_i, Y_i)\}_{i=1}^{|\mathcal{D}|}$, where $X_i$ represents a slide and $Y_i={1,2,\cdots,n}$ represents its corresponding label. Our goal is to establish a mapping from the $\mathcal{X}$ domain to the $\mathcal{Y}$ domain. However, learning direct mapping is extremely challenging due to the gigapixel nature of WSI. Thus, each slide is commonly segmented into patches $X_i=\{x_{i,k}\}_{k=1}^{N}$ for multiple instance learning, where $x_{i,k}$ denotes $k$-th patch cut from $i$-th slide, the number $N$ of patches in a slide is often different from others, and patch's corresponding label $y_{i,k}$ remains unknown during training. In the binary MIL classification problem, the relationship between patch-level labels $y_{i,k}$ and bag-level labels $Y_i$ typically follows the following pattern:
\begin{equation}\label{prin}
Y_i = \begin{cases}
    +1, &\text{ $\exists k:  y_{i,k} = +1$},\\
    -1, &\text{ $\forall k:  y_{i,k} = -1$},
\end{cases}
\end{equation}
where $+1$ represents the positive sample and $-1$ represents the negative sample. 

Multiple instance learning involves a three-step modeling process: (1) Instances are transformed into low-dimensional embeddings through a feature encoder:
\begin{equation}\label{equ2}
h_{i,k}=f_\omega(x_{i,k}).
\end{equation}
(2) All instance embeddings are aggregated into a bag-level representation:
\begin{equation}\label{equ3}
H_i=g_\varphi(h_{i,1}, h_{i,2}, \cdots, h_{i,N}),
\end{equation}
where the attention mechanism is often used for aggregation:
\begin{equation}
H_i = \sum_{k=1}^{n} a_{i,k} h_{i,k},
\end{equation}
where $a_{i,k} = \sigma(h_{i,k})$ represents the attention value of patch $x_{i,k}$. In the case of ABMIL, a gated attention mechanism is adopted as follows:
\begin{equation}
\sigma(h_{i,k}) = \frac{\exp\left\{w^T \left(\tanh(V_1 h_{i,k}) \odot sigm(V_2 h_{i,k})\right)\right\}}{\sum_{j=1}^{N} \exp\left\{w^T \left(\tanh(V_1 h_{i,j}) \odot sigm(V_2 h_{i,j})\right)\right\}}.
\end{equation}
(3) Predict the aggregated bag-level representation with an MLP as follows:
\begin{equation}\label{equ4}
\hat{Y_i}=\arg\max \Gamma_\phi(H_i).
\end{equation}

\subsection{Heuristic clustering strategy}

Task-oriented feature fine-tuning methods are usually based on the idea of distillation which is typically iterative. Take the first round of iteration as an example: (1) extract features with an encoder initialized with pre-trained weights $\omega_{(0)}$ which are trained usually on ImageNet-1K as Eq. \ref{equ2}; (2) train the aggregator $g_{\varphi}$ and MLP $\Gamma{\phi}$ following Eq. \ref{equ3} and Eq. \ref{equ4}; (3) distill the patch-level information from the $g_{\varphi_{(0)}^*}$ and $\Gamma_{\phi_{(0)}^*}$; (4) fine-tune the feature encoder $f_\omega$ by the patch-level information and update the features with the new encoder $f_{\omega_{(1)}^*}$. Subsequently, the aforementioned steps can be repeatedly executed until reaching an optimal outcome or satisfying the termination criteria.

Taking the framework of task-oriented feature fine-tuning methods, a common distillation practice for feature encoder updating is the pseudo label strategy. In this context, a patch-level pseudo label $z_{i,k}$ is assigned to each selected instance $x_{i,k}$. This supervised information is employed to update the feature encoder $f_\omega$, making the selection of pseudo labels crucial. As learning too many noisy samples could devastate the fine-tuning process, we first introduce a pseudo label cleaning module. Furthermore, overly simplistic samples distilled may fail to provide the feature encoder with ample information, resulting in overfitting issues. Hence, we introduce a hard negative sample mining module, designed to furnish the training process with adversarial information, thereby enhancing the generalization ability. 

\begin{figure*}[ht!]
    \centering
    \includegraphics[width=\linewidth]{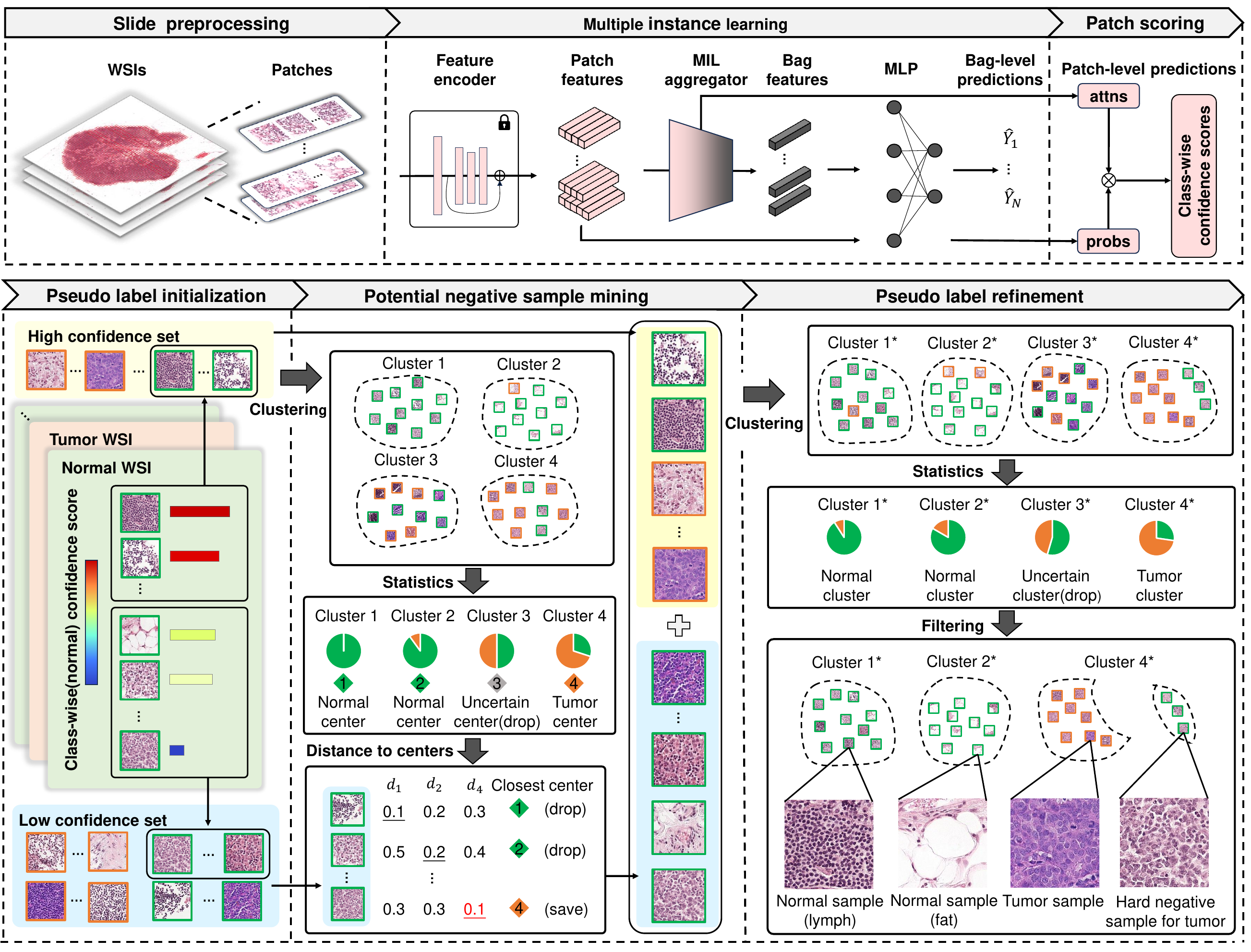}
    \caption{Overview of the proposed heuristic clustering-driven feature fine-tuning method. After slide preprocessing and feature extraction, we can obtain a well-trained MIL model by weakly supervised learning. Then, we freeze the MIL aggregator to get class-wise confidence scores, which split all patches into high and low confidence sets. Patches in different sets are assigned with different pseudo labels. We use the first heuristic clustering for potential negative sample mining and the second heuristic clustering for positive sample cleaning and hard negative sample searching. With this pseudo label refinement, a patch-level dataset with pure positive samples and hard negative samples is constructed for feature encoder fine-tuning.}
    \label{main}
\end{figure*}
\begin{algorithm}[t!]
\KwIn{clusters set $\{cluster_i\}_{j=1}^{C}$, label $a$, threshold $\theta$}
\KwResult{the cluster set $L_a$ for label $a$}

\For{$j \gets 1$ \KwTo $C$}{
        $count = 0, total = 0$\;
        \For{$(x_{i,k},z_{i,k}) \in cluster_j$}{
            \If{$ z_{i,k}=a$}{
            $count \gets count + 1$ \;}
            $total \gets total + 1$ \;
            }
        $p_j \gets count \div total$\;
    }
\eIf {$\max p_j \leq \theta$}{
    $L_a \gets \{cluster_{\arg\max p_j}\}$}
    {$L_a \gets \{cluster_j|p_j > \theta\}$
}
\caption{Clusters classification (for single class)}
\label{alg1}
\end{algorithm}

Then we will introduce each component of the heuristic clustering-driven feature fine-tuning method (HC-FT) in Fig. \ref{main}. 
\subsubsection{Pseudo label initialization}
In attention-based multiple instance learning, assigning each selected patch with the bag-level label based on the attention score seems a logical choice. However, as the attention score fails to reflect category information, we adopt a class-wise confidence-based pseudo label initialization as follows: (1) Given an optimal MIL model with $f_{\omega^{(0)}}$, $g_{\varphi^{(0)}}$ and $\Gamma_{\phi^{(0)}}$, the attention score set $A_i=\{a_{i,1}, a_{i,2}, \cdots, a_{i,N}\}$ and instance probability set $P_i=\{p_{i,1}, p_{i,2}, \cdots, p_{i,N}\}$ for each slide can be obtained, where $p_{i,k} = \Gamma_{\phi}(h_{i,k})$. Then a class-wise confidence score $s_{i,k}$ can be assigned to each patch $x_{i,k} \in X_i$ as:
\begin{equation}
s_{i,k} = a_{i,k}^{Y_i} \cdot p_{i,k}^{Y_i},
\end{equation}
where $p_{i,k}^{Y_i}$ denotes the class-wise probability, sharing the same dimension with $a_{i,k}$. (2) By ranking the class-wise confidence score set $S_i=\{s_{i,1},s_{i,2},\cdots,s_{i,N}\}$ in descending order for each slide, an ordered index set $M_i = \{m_{i,1}, m_{i,2}, \cdots, m_{i,N}\}$ is obtained. (3) Assign patches with top confidences pseudo labels:
\begin{equation}
z_{i,k}= \begin{cases}
    Y_i, &\text{$1 \leq m_{i,n} \leq K_{t}$},\\
    -1, &\text{$m_{i,n} > K_{t}$}.
\end{cases}
\end{equation}
where $K_{t}$ represents the number of patches with highest $K_{t}$ confidence scores. Notably, $K_{t}$ increases with each iteration $t$, following the dynamic adjustment scheme:
\begin{equation}
 K_{t} = \min ( (t + 1) \cdot K_0 \cdot log_{10}(N), \frac{N}{3}),
\end{equation}
where $K_0$ is a hyper-parameter determined through experiments. Intuitively, the base number $K_t$ should dynamically change to accommodate bags containing varying numbers of patches. All patches are thus allocated into a high confidence score set $T_h=\{(x_{i,k},z_{i,k}) | z_{i,k}=Y_i\}$ and a low confidence score set $T_l=\{(x_{i,k},z_{i,k}) | z_{i,k}=-1\}$. 
\subsubsection{Potential negative sample mining}
Initially, clustering is conducted on all patches from the high-confidence set $T_h$. The category of each cluster center is assigned based on the pseudo label content. As shown in Alg. \ref{alg1}, we first apply K-Means \citep{Kmeans} to get $C$ clusters. In each cluster, we count the number of pseudo labels for each category, then assign the category pseudo label with the highest number to each cluster. We also introduce a threshold $\theta$ to eliminate those chaotic clusters.

For one class $a$, we can obtain a set of clusters $L_a$, where cluster centers represent centers of the class $a$. Utilizing all class centers, we can mine more negative samples within the low confidence group $T_l$. Specifically, for the cluster set $L=\{L_1,\cdots,L_{n}\}$, the distance $d_{{x_{i,k}},L_a}$ between each patch in $T_l$ and class center can be calculated as:
\begin{equation}
d_{{x_{i,k}},L_a} = \|h_{i,k}-L_a\|.
\end{equation}

The new pseudo label is assigned to each patch in $T_l$ by:
\begin{equation}
z_{i,k} =
\begin{cases}
n + \arg\min_j d_{{x_{i,k}},L_a} & \text{ if } j > Y_i,\\
-1 & \text{ if } j \leq Y_i.
\end{cases}
\end{equation}
\begin{algorithm}[t!]
\caption{Heuristic clustering strategy for feature fine-tuning}
\label{heuristic}
\KwIn{high confidence patch set $T_h$ and low confidence patch set $T_l$}
\KwResult{patch-level dataset with pseudo labels $\mathcal{D}^*$}
\tcp{pseudo label initialization}
\For{$x_{i,k} \in T_h$}{
    $z_{i,k} \gets Y_i$ 
}
\For{$x_{i,k} \in T_l$}{
    $z_{i,k} \gets -1$ 
}
\tcp{potential negative sample mining}
$\text{clusters}\gets \text{run K-means in }T_h$\;
$L=\{L_j\}_{j=1}^{n}\gets\text{run Algorithm 1 in clusters} $\;

\For{$x_{i,k} \in T_l$}{
    \For{$a \gets 1$ \KwTo $n$}{
        $d_{x_{i,k},L_a} = L_2(x_{i,k}, center_{L_a})$\;
    }
    $z_{i,k} \gets C + l$ s.t. $\underset{j}{\arg\min}d_{x_{i,k},L_a} > Y_i$ \;
}
$N \gets T_l \setminus \{(x_{i,k},z_{i,k}) \mid z_{i,k} \ne -1\}$\;
$\text{new clusters}\gets \text{run K-means in }N + T_h$\;
$L^*=\{L_j^*\}_{j=1}^{n} \gets $ run Algorithm 1 in new clusters\;
\tcp{pseudo label refinement}
\For{$x_{i,k} \in N$}{
    \For{$a \gets 1$ \KwTo $n$}{
$N \gets N \setminus \{(x_{i,k},z_{i,k})|x_{i,k} \in L_{a\leq Y_i}^*\}$\;
    }
}
\For{$x_{i,k} \in T_h$}{
    $T_h \gets T_h \setminus \{(x_{i,k},z_{i,k})|x_{i,k} \notin L_{Y_i}^*\}$ \;
    \For{$a \gets 1$ \KwTo $n$}{
        $N \gets N \cup \{(x_{i,k},z_{i,k})|x_{i,k} \in L_{a>Y_i}^*\}$\;
    }
}
$\mathcal{D}^* \gets T_h \cup N$ \;

\end{algorithm}
The WSI datasets have an inherent character in that the bag-level label is determined by the labels of the patches with the most severe lesions. In other words, all patch-level labels should be less than or equal to the bag-level label in a slide:
\begin{equation}\label{tui}
y_{i,k}\leq Y_i.
\end{equation}
One patch that belongs to $j$-th label cluster set $L_j$ while $j $ is bigger than its bag-level label $Y_i$ is a potential negative sample, these patches are allocated in a set $N_{original}$:
\begin{equation}\label{negative}
    N_{original} = \{x_{i,k} \in L_{j > Y_i}| x_{i,k} \in T_l\}_{j=1}^{n}.
\end{equation}
With Eq. \ref{tui}, we can ensure the patches we mined are negative samples. We append a new category for these negative samples, hoping that the model will pay increased attention to them. For clarity, we summarize the process of heuristic clustering strategy in Alg. \ref{heuristic}.

\subsubsection{Pseudo label refinement}
We integrate set $T_h$ and set $N_{original}$ as a new set, and repeat Alg. \ref{alg1} to analyze the status of each new cluster.

\textbf{A. \textit{Hard negative sample searching}}

Hard negative samples refer to negative samples that are difficult for the model to classify. We propose a hard sample mining module to identify hard negative samples for the model to learn, thereby enhancing the distinction ability of the encoder.

After the re-clustering process, a new set $N_{middle_l}$ appears with filtering the $N_{original}$ as below:
\begin{equation}
N_{middle_l} = N_{original} / \{x_{i,k} \in L_{j \leq Y_i} | x_{i,k} \in N_{original}\}_{j=1}^n.
\end{equation}
The objective of re-clustering is to ascertain whether those identified as hard samples are still challenging to distinguish. The set $N_{middle_l}$ is part of the final hard negative sample set, and the remaining portion originates from the high confidence set $T_h$. Following the same step in Eq. \ref{negative} in high confidence set $T_h$, we can get a hard negative samples set $N_{middle\_h}$ as:
\begin{equation}
N_{middle\_h} = \{x_{i,k} \in L_{j>Y_i}|x_{i,k} \in T_h\}_{j=1}^{n}.
\end{equation}

Combined the two hard negative samples sets $N_{middle_h}$ and $N_{middle_l}$, we can obtain:
\begin{equation}
N_{final} = N_{middle\_h} + N_{middle\_l},
\end{equation}
where $N_{final}$ is the final hard negative sample set. In addition to containing negative samples with high attention scores, this set includes representative negative samples that are ignored by attention sorting, which provides more adversarial information for encoder fine-tuning.

\textbf{B. \textit{Positive samples cleaning}}

The classification performance of the model mainly depends on the quality of positive samples. Incorrect pseudo labels can significantly cause the model to converge to a bad solution. Therefore, we undertake a cleaning process for pseudo labels derived from confidence scores. To be specific, given a center set $L_j$ and a high confidence set $T_h$, patches whose pseudo labels align with the category represented by their cluster re-clustering are considered to be more accurate, Thus a filtering process is performed as below:
\begin{equation}
T_h = T_h / \{x_{i,k} \in I_{j \ne Y^i}|x_{i,k} \in T_h\}_{j=1}^n.
\end{equation}

\subsection{Task-oriented feature fine-tuning}

Through the two aforementioned modules, we can obtain a patch-level dataset $\mathcal{D}^* = \{(x_{i,k},z_{i,k})\}$ annotated with relatively clean pseudo labels for feature fine-tuning. Notably, the patch-level classification task typically encompasses $2n-1$ classes, where initial $n$ classes consist of positive samples from each category, and the remaining $n-1$ classes represent the corresponding hard negative samples, excluding the non-tumor class.

\section{Experiment}
\subsection{Datasets and metrics}
Our experiments employ two publicly available whole slide image classification datasets, namely CAMELYON16 \citep{camelyon16} and BRACS \citep{brancati2021bracs}.

CAMELYON16 dataset is a binary classification dataset focused on the detection of early-stage breast cancer lymph node metastases. It consists of 399 hematoxylin and eosin (H\&E) stained WSIs, sourced from two distinct medical institutions, with 270 allocated for training and 129 for testing. Additionally, this dataset provides detailed pixel-level manual annotations of metastatic tumor areas, thereby facilitating in-depth patch-level analysis and enhanced visualization capabilities. We conducted a three-fold cross-validation, where 270 training samples were divided into a training and a validation set at a ratio of 8:2. All slides were cut into non-overlapping patches of $256 \times 256$ pixels at a $20 \times$ magnification.
\begin{table*}[t!]
\centering
\caption{Slide-level performance with various feature extraction techniques on CAMELYON16 and BRACS datasets. The yellow boxes are performances of our methods, the best results are highlighted and the second best results are underlined.}
\label{tab1}
\definecolor{lightyellow}{rgb}{1.0, 1.0, 0.88}
\definecolor{lightgreen}{HTML}{3F8061}
\resizebox{\textwidth}{!}{
\begin{tabular}{@{}c|l|lll|lll|lll@{}}
\toprule
\multirow{2}{*}{Methods} & \multirow{2}{*}{Encoder weight} & \multicolumn{3}{c|}{CAMELYON16} & \multicolumn{3}{c|}{BRACS} & \multicolumn{3}{c}{Average (\%)} \\
                       &          & ACC (\%)     & AUC (\%)     & F1 (\%)      & ACC (\%)    & AUC (\%)    & F1 (\%)     & ACC  & AUC & F1 \\ 
\midrule
\multirow{9}{*}{\makecell[c]{ABMIL \\ \citep{ABMIL}}} & w/Random   & $62.79_{\pm1.09}$  & $58.20_{\pm0.91}$  & $49.17_{\pm2.39}$  & $39.85_{\pm4.34}$ &           
                       $62.27_{\pm1.22}$ & $25.26_{\pm7.75}$ & $51.32$    & $60.25$   & $37.22$  \\
                       & w/ImageNet-1K & $81.91_{\pm0.36}$  & $78.85_{\pm2.00}$  & $80.71_{\pm0.64}$  & $50.57_{\pm7.08}$ & $72.42_{\pm3.09}$ & $44.31_{\pm7.54}$ & $66.24$    & $75.63$   & $62.51$  \\
                       & w/Barlow Twins & $68.73_{\pm4.02}$  & $79.59_{\pm3.13}$  & $69.16_{\pm3.94}$  & $62.07_{\pm0.94}$ & $81.97_{\pm0.89}$ & $59.36_{\pm3.60}$ & $65.40$    & $80.78$   & $64.26$  \\
                       & w/MoCo V2   & $71.84_{\pm12.08}$ & $83.23_{\pm1.61}$  & $71.28_{\pm12.58}$ & $59.77_{\pm3.39}$ & $76.67_{\pm2.90}$ & $53.78_{\pm4.65}$ & $65.80$    & $79.95$   & $62.53$  \\
                       & w/SWAV     & $81.62_{\pm1.32}$  & $84.70_{\pm2.53}$  & $81.61_{\pm1.15}$  & \underline{$64.75_{\pm0.54}$} & $84.25_{\pm0.17}$ & \underline{$64.19_{\pm0.28}$} & $73.20$    & $84.47$   & $72.90$  \\
                       & w/WSI-FT   & $77.52_{\pm6.70}$  & $90.99_{\pm1.27}$  & $77.57_{\pm6.91}$  & $58.62_{\pm1.88}$ & $77.68_{\pm0.99}$ & $55.38_{\pm2.55}$ & $68.07$    & $84.33$   & $66.48$  \\
                       & w/BCL      & \underline{$89.15_{\pm1.90}$}  & \underline{$94.34_{\pm1.81}$}  & \underline{$89.12_{\pm1.96}$}  & $64.37_{\pm1.88}$ & \underline{$84.67_{\pm0.51}$} & $63.39_{\pm2.35}$ & \underline{$76.76$}    & \underline{$89.51$}   & \underline{$76.26$}  \\
                       & \textbf{w/Ours}\cellcolor{lightyellow} & $\mathbf{92.77_{\pm0.73}}$\cellcolor{lightyellow} & $\mathbf{97.13_{\pm1.18}}$\cellcolor{lightyellow} & $\mathbf{92.27_{\pm0.70}}$\cellcolor{lightyellow} & $\mathbf{67.82_{\pm2.81}}$\cellcolor{lightyellow} & $\mathbf{85.85_{\pm0.56}}$\cellcolor{lightyellow} & $\mathbf{65.37_{\pm2.63}}$\cellcolor{lightyellow} & 
                       $\mathbf{80.29}$\cellcolor{lightyellow} & $\mathbf{91.49}$\cellcolor{lightyellow} & $\mathbf{78.82}$\cellcolor{lightyellow} \\ 
                       & $\Delta$ \textbf{ over others} & \textcolor{lightgreen}{$+3.62$} & \textcolor{lightgreen}{$+2.79$} & \textcolor{lightgreen}{$+3.15$} & \textcolor{lightgreen}{$+3.07$} & \textcolor{lightgreen}{$+1.18$} & \textcolor{lightgreen}{$+1.18$} & \textcolor{lightgreen}{$+3.53$} & \textcolor{lightgreen}{$+1.98$} & \textcolor{lightgreen}{$+2.56$} \\
\midrule
\midrule
\multirow{9}{*}{\makecell[c]{CLAM \\ \citep{clam}}}  & w/Random   & $53.75_{\pm11.70}$ & $43.74_{\pm8.85}$  & $38.52_{\pm12.67}$ & $36.78_{\pm0.00}$ & $52.75_{\pm0.00}$ & $19.78_{\pm0.00}$ & 45.27    & 48.25   & 29.15  \\
                       & w/ImageNet-1K & $83.72_{\pm1.27}$  & $79.62_{\pm6.26}$  & $81.27_{\pm1.91}$  & $55.94_{\pm1.95}$ & $73.91_{\pm1.74}$ & $49.24_{\pm2.53}$ & 69.83    & 76.77   & 65.26  \\
                       & w/Barlow Twins & $73.39_{\pm12.44}$ & $79.93_{\pm6.26}$  & $81.27_{\pm1.91}$  & $61.69_{\pm3.02}$ & $77.76_{\pm3.63}$ & $54.76_{\pm4.89}$ & 67.54    & 78.85   & 63.99  \\
                       & w/MoCo V2   & $72.10_{\pm6.85}$  & $80.35_{\pm2.35}$  & $72.28_{\pm6.69}$  & $63.60_{\pm3.02}$ & $83.26_{\pm1.74}$ & $60.16_{\pm3.16}$ & 67.85    & 81.81   & 66.22  \\
                       & w/SWAV     & $79.85_{\pm12.70}$ & $92.97_{\pm2.77}$  & $79.53_{\pm13.14}$ & \underline{$64.37_{\pm2.48}$} & \underline{$83.91_{\pm1.32}$} & \underline{$64.45_{\pm1.99}$} & 72.11    & \underline{88.44}   & 71.99  \\
                       & w/WSI-FT   & $82.95_{\pm1.68}$  & $89.46_{\pm1.54}$  & $82.88_{\pm1.54}$  & $62.07_{\pm2.82}$ & $80.31_{\pm0.18}$ & $60.96_{\pm3.21}$ & 72.51    & 84.88   & 71.92  \\
                       & w/BCL      & \underline{$88.63_{\pm0.96}$}  & \underline{$92.78_{\pm3.22}$}  & \underline{$88.54_{\pm1.15}$}  & $62.84_{\pm1.43}$ & $83.23_{\pm1.48}$ & $58.93_{\pm1.93}$ & \underline{75.73}    & 88.01   & \underline{73.74}  \\
                       & \textbf{w/Ours}\cellcolor{lightyellow} & $\mathbf{91.47_{\pm1.68}}$\cellcolor{lightyellow} & $\mathbf{95.25_{\pm2.71}}$\cellcolor{lightyellow} & $\mathbf{91.42_{\pm1.66}}$\cellcolor{lightyellow} & $\mathbf{68.20_{\pm1.08}}$\cellcolor{lightyellow} & $\mathbf{84.91_{\pm0.60}}$\cellcolor{lightyellow} & $\mathbf{64.49_{\pm0.82}}$\cellcolor{lightyellow} & $\mathbf{79.83}$\cellcolor{lightyellow} & $\mathbf{90.08}$\cellcolor{lightyellow} & $\mathbf{77.96}$\cellcolor{lightyellow} \\
                       & $\Delta$ \textbf{ over others} & \textcolor{lightgreen}{+2.84} & \textcolor{lightgreen}{+2.28} & \textcolor{lightgreen}{+2.88} & \textcolor{lightgreen}{+3.83} & \textcolor{lightgreen}{+1.00} & \textcolor{lightgreen}{+0.04} & \textcolor{lightgreen}{+4.10} & \textcolor{lightgreen}{+1.64} & \textcolor{lightgreen}{+4.22} \\
\midrule
\midrule
\multirow{9}{*}{\makecell[c]{DSMIL \\ \citep{DSMIL}}} & w/Random   & $64.83_{\pm3.44}$  & $59.88_{\pm4.51}$  & $48.06_{\pm0.82}$  & $36.78_{\pm0.00}$ & $62.19_{\pm1.30}$ & $20.46_{\pm0.96}$ & 50.80    & 61.03   & 34.26  \\
                       & w/ImageNet-1K & $76.48_{\pm2.56}$  & $76.89_{\pm7.29}$  & $75.34_{\pm2.79}$  & $53.26_{\pm6.25}$ & $72.10_{\pm3.95}$ & $51.25_{\pm8.36}$ & 64.87    & 74.50   & 63.30  \\
                       & w/Barlow Twins       & $68.47_{\pm5.71}$  & $61.00_{\pm9.94}$     & $63.55_{\pm10.07}$ & $55.94_{\pm3.02}$ & $78.59_{\pm1.26}$ & $50.71_{\pm0.88}$ & 62.21    & 69.80   & 57.13  \\
                       & w/MoCo V2   & $56.07_{\pm6.77}$  & $53.96_{\pm7.17}$  & $53.61_{\pm6.22}$  & $65.14_{\pm1.08}$ & $81.26_{\pm0.87}$ & $62.62_{\pm0.83}$ & 60.61    & 67.61   & 58.11  \\
                       & w/SWAV     & $76.23_{\pm8.01}$  & $80.04_{\pm15.04}$ & $76.08_{\pm8.31}$  &\underline{$65.14_{\pm2.36}$} & $82.01_{\pm1.21}$ & \underline{$63.67_{\pm1.73}$} & 70.68    & 81.02   & 69.88  \\
                       & w/WSI-FT   & $83.20_{\pm0.73}$  & $91.57_{\pm1.51}$  & $83.17_{\pm0.66}$  & $57.09_{\pm0.54}$ & $77.08_{\pm0.26}$ & $51.71_{\pm2.08}$ & 70.15    & 84.21   & 67.44  \\
                       & w/BCL      & \underline{$90.18_{\pm3.12}$}  & \underline{$92.81_{\pm2.39}$}  & \underline{$90.12_{\pm3.11}$}  & $60.15_{\pm4.34}$ & \underline{$82.61_{\pm1.87}$} & $57.18_{\pm4.87}$ & \underline{75.17}    & \underline{87.71}   & \underline{73.65}  \\
                       & \textbf{w/Ours} \cellcolor{lightyellow} & $\mathbf{91.21_{\pm1.31}}$\cellcolor{lightyellow} & $\mathbf{95.57_{\pm2.44}}$\cellcolor{lightyellow} & $\mathbf{91.17_{\pm1.26}}$\cellcolor{lightyellow} & $\mathbf{67.05_{\pm2.36}}$\cellcolor{lightyellow} & $\mathbf{84.32_{\pm0.91}}$\cellcolor{lightyellow} & $\mathbf{64.42_{\pm3.72}}$\cellcolor{lightyellow} & $\mathbf{79.13}$\cellcolor{lightyellow} & $\mathbf{89.95}$\cellcolor{lightyellow} & $\mathbf{77.79}$\cellcolor{lightyellow} \\
                       & $\Delta$ \textbf{ over others} & \textcolor{lightgreen}{+1.03} & \textcolor{lightgreen}{+2.76} & \textcolor{lightgreen}{+1.05} & \textcolor{lightgreen}{+1.91} & \textcolor{lightgreen}{+1.71}  & \textcolor{lightgreen}{+0.75} & \textcolor{lightgreen}{+3.96}  & \textcolor{lightgreen}{+2.24} & \textcolor{lightgreen}{+4.14} \\
\bottomrule
\end{tabular}
}
\end{table*}
\begin{table*}[ht!]
\centering
\caption{Slide-level performance with various MIL methods on CAMELYON16 and BRACS datasets. The yellow boxes are performances of our methods, the best results are highlighted and the second best results are underlined.}
\label{tab2}
\begin{tabular}{@{}cccccccccc@{}}
\toprule
\multirow{2}{*}{Methods} & \multicolumn{3}{c}{CAMELYON16} & \multicolumn{3}{c}{BRACS} & \multicolumn{3}{c}{Average (\%)} \\
        &  ACC (\%)     & AUC (\%)     & F1 (\%)      & ACC (\%)    & AUC (\%)    & F1 (\%)     & ACC  & AUC & F1 \\ 
\midrule
ABMIL    & $81.91_{\pm0.36}$ & $78.85_{\pm2.00}$ & $80.71_{\pm0.64}$ & $50.57_{\pm7.08}$ & $72.42_{\pm3.09}$ & $44.31_{\pm7.54}$ & $66.24$    & $75.63$   & $62.51$  \\
CLAM\_SB & $83.72_{\pm1.27}$ & $79.62_{\pm6.26}$ & $81.27_{\pm1.91}$ & $55.94_{\pm1.95}$ & $73.91_{\pm1.74}$ & $49.24_{\pm2.53}$ & $69.83$    & $76.77$   & $65.26$  \\
CLAM\_MB & $83.21_{\pm2.03}$ & $80.75_{\pm2.43}$ & $82.22_{\pm2.29}$ & $55.17_{\pm3.76}$ & $74.33_{\pm2.79}$ & $48.98_{\pm6.32}$ & $69.19$    & $77.54$   & $65.60$  \\
DSMIL    & $76.48_{\pm2.56}$ & $76.89_{\pm7.29}$ & $75.34_{\pm2.79}$ & $53.26_{\pm6.25}$ & $72.10_{\pm3.95}$ & $51.25_{\pm8.36}$ & $64.87$    & $74.50$   & $63.30$  \\
ACMIL    & $81.91_{\pm4.75}$ & $83.37_{\pm6.91}$ & $80.29_{\pm5.37}$ & $55.17_{\pm0.94}$ & $73.57_{\pm0.77}$ & $42.34_{\pm0.64}$ & $68.54$    & $78.47$   & $61.32$  \\
TransMIL & $85.27_{\pm1.27}$ & $90.91_{\pm1.39}$ & $83.52_{\pm1.42}$ & $60.15_{\pm2.87}$ & $76.00_{\pm0.15}$ & $51.31_{\pm2.51}$ & $72.71$    & $83.45$   & $67.42$  \\
DTFD     & $85.53_{\pm1.46}$ & $84.17_{\pm3.88}$ & $84.95_{\pm1.69}$ & $59.39_{\pm4.44}$ & $76.54_{\pm1.90}$ & $58.64_{\pm5.68}$ & $72.46$    & $80.36$   & $71.80$  \\
WSI-FT   & $77.52_{\pm6.70}$ & $90.99_{\pm1.27}$ & $77.57_{\pm6.91}$ & $58.62_{\pm1.88}$ & $77.68_{\pm0.99}$ & $55.38_{\pm2.55}$ & $68.07$    & $84.33$   & $66.48$  \\
BCL      & \underline{$89.15_{\pm1.90}$} & \underline{$94.34_{\pm1.81}$} & \underline{$89.12_{\pm1.96}$} & \underline{$64.37_{\pm1.88}$} & \underline{$84.67_{\pm0.51}$} & \underline{$63.39_{\pm2.35}$} & \underline{$76.76$}    & \underline{$89.51$}   & \underline{$76.26$}  \\
Ours\cellcolor{lightyellow} & $\mathbf{92.77_{\pm0.73}}$\cellcolor{lightyellow} & $\mathbf{97.13_{\pm1.18}}$\cellcolor{lightyellow}  & $\mathbf{92.27_{\pm0.70}}$\cellcolor{lightyellow} & $\mathbf{67.82_{\pm2.81}}$\cellcolor{lightyellow} & $\mathbf{85.85_{\pm0.56}}$\cellcolor{lightyellow} & $\mathbf{65.37_{\pm2.63}}$\cellcolor{lightyellow} & $\mathbf{80.29}$\cellcolor{lightyellow} & $\mathbf{91.49}$\cellcolor{lightyellow} & $\mathbf{78.82}$\cellcolor{lightyellow} \\
\bottomrule
\end{tabular}
\end{table*}

BReAst Carcinoma Subtyping (BRACS) dataset emerges as a promotion for the automated detection and classification of breast tumors, also constituted by H\&E stained histopathological images. The dataset encompasses 547 whole slide images. Each WSI has been meticulously annotated according to consensus by a panel of three board-certified pathologists, categorizing different types of lesions. Specifically, this dataset includes three types of lesions: benign, malignant, and atypical, further divided into seven subtypes. In this paper, we focus on the three-class classification. We adhered to the official division and employed distinct seeds across three experimental trials. All slides were cut into non-overlapping patches of $256 \times 256$ pixels at a $5 \times$ magnification.

In our experiments, we employ several evaluation metrics in bag-level classification, including accuracy (ACC), the area under the receiver operating characteristic curve (AUC), and the harmonic mean of precision and recall (F1 score). For the multi-class dataset BRACS, AUC is computed using a one-versus-all mode. In addition, we also conducted a patch-level evaluation with another two evaluation metrics: the free-response receiver operating characteristic (FROC) \citep{Froc} and the competition performance metric (CPM) score \citep{CPM}. FROC is defined as the plot of sensitivity versus the average number of false positive patches per image (FPI) and CPM is defined as the average sensitivity at seven predefined false positive rates which seven values are 0.125, 0.25, 0.5, 1, 2, 4, and 8 in FROC curve and was calculated as:
\begin{equation}
    \text{CPM} = \frac{1}{N} \sum_{i=\{0.125, 0.25, 0.5, 1, 2, 4, 8\}} \text{Sensitivity}_{\text{FPI}=i}.
\end{equation}

\subsection{Implementation details}

All experiments were conducted on one NVIDIA RTX 3090 GPU which possesses 24 GB memory. Our methodology was implemented within the PyTorch v2.0 deep learning framework. In our experiments, we applied the resnet50 \citep{he2015deep} backbone as the feature encoder, and Adam optimizer was utilized for training. For the MIL training, the batch size was set to 1, with an initial and minimum learning rate of $1 \times 10^{-3}$ and $1 \times 10^{-4}$, respectively. For the encoder fine-tuning, the batch size was set to 64, with a fixed learning rate of $5 \times 10^{-5}$. The maximum number of epochs for training both the MIL training and the encoder fine-tuning was established at 200. The early stopping strategy was implemented to prevent overfitting, which was monitored through the loss on the validation set.

\subsection{Methods for comparison}
In general, we evaluate the performance of different features under certain MIL models and the performance of certain features under different MIL models.

As our method aims at enhancing feature representation, we first evaluate features generated by different approaches: (1) Random initialization. (2) Pre-trained on natural images, Imagenet-1K \citep{he2015deep}. (3) The weight from employing self-supervised methods on extensive WSI datasets, including Barlow Twins \citep{BT}, MoCo V2 \citep{mocov2}, and SWAV \citep{swav}. (4) Task-oriented feature fine-tuning methods, including WSI-FT \citep{WSIFT} and BCL \citep{BCL}.
\begin{figure*}[ht!]
    \centering
    \includegraphics[width=0.85\linewidth]{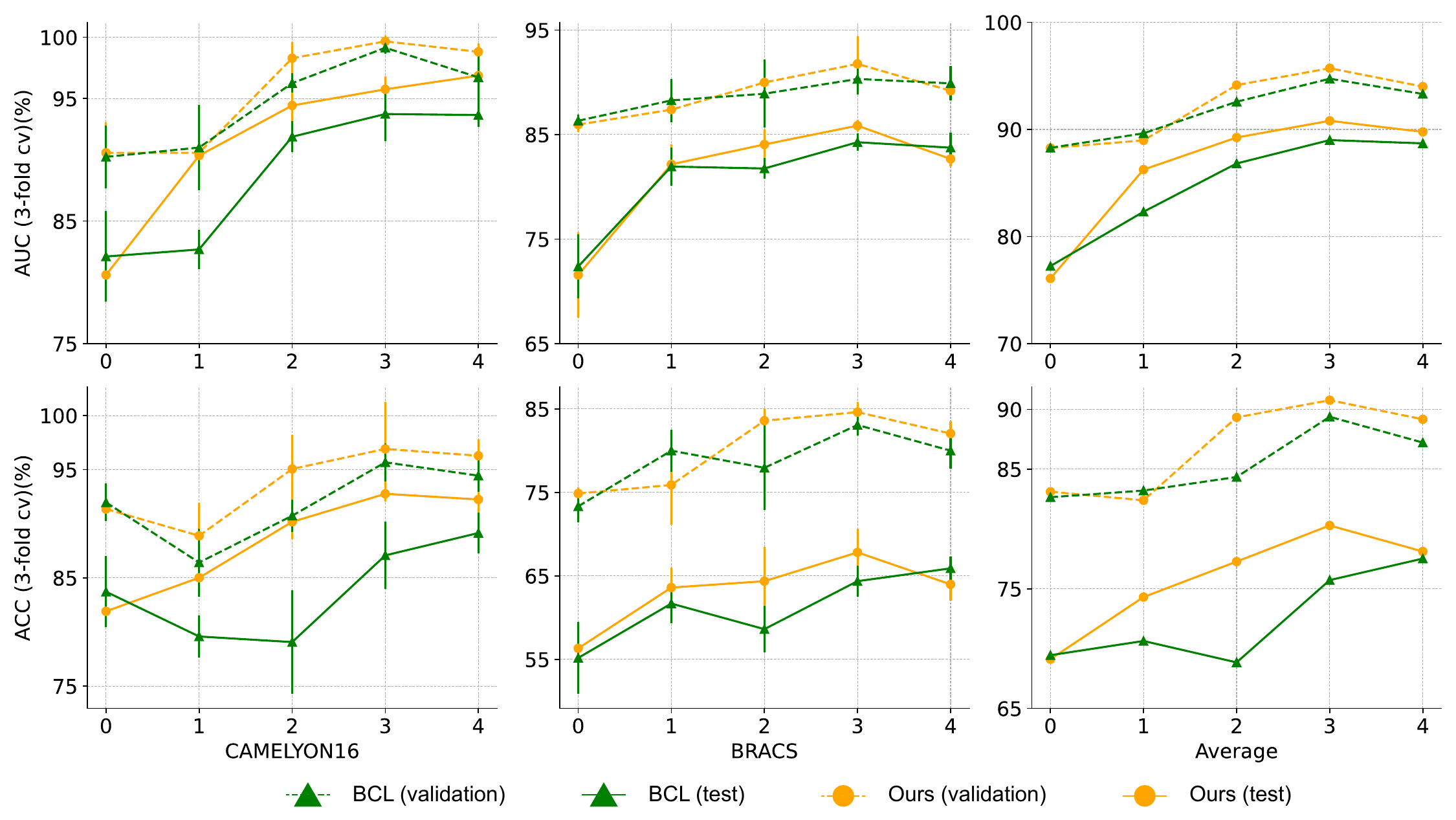}
    \caption{Validation and test performance variations across different training iterations on CAMELYON16 and BRACS datasets.}
    \label{figure3}
\end{figure*}

Additionally, as a multi-instance learning approach for WSI classification, we also evaluate other relevant MIL methods: (1) The classic ABMIL \citep{ABMIL}. (2) Four variants of ABMIL, including single-attention-branch CLAM\_SB \citep{clam}, multi-attention-branch CLAM\_MB \citep{clam}, non-local attention pooling DSMIL \citep{DSMIL} and muti-head attention ACMIL \citep{ACMIL}. (3) Transformer-based MIL, TransMIL \citep{TransMIL}. (6) Double-tier feature distillation MIL, DTFD \citep{DTFD}. (7) End-to-end feature encoder fine-tuning, WSI-FT \citep{WSIFT}. (8) Bayesian collaborative learning-based MIL, BCL \citep{BCL}. All results of these methods are conducted using their official codes under the same settings.
\section{Results}
\subsection{Quantitative comparison}

To validate the efficiency of the embedding generated by our approach, we conducted a comparative analysis of performance results obtained with various feature extraction techniques, as reported in Tab. \ref{tab1}. Initially, features derived from a randomly initialized feature encoder exhibited limited classification ability. Through supervised learning with ImageNet-1K, significant enhancements were observed in the representational capacity of embeddings, especially in the CAMELYON16 dataset. Self-supervision based methods facilitated fine-tuning of the feature encoder using extensive pathological images. Embeddings generated through Barlow Twins, MoCo V2, and SWAV failed to demonstrate superior performance in the CAMELYON16 dataset but exhibited improved performance in the BRACS dataset. This phenomenon could be attributed to the dataset's inherent characteristics. The utilization of self-supervised weights may inadvertently complicate features, resulting in different MIL classification outcomes according to the category distinctions in different datasets. In task-oriented feature fine-tuning methods, BCL outperformed most of the other methods as it distills efficient patch-level information from MIL models for encoder fine-tuning. Notably, our method consistently outperformed other feature fine-tuning approaches in both datasets. This robust performance underscores the effectiveness and versatility of our approach. Furthermore, our results remained optimal even when applied to different MIL models, highlighting the generalizability of our approach.
\begin{table}[t!]
\centering
\caption{Patch-level performance with various MIL methods on the CAMELYON16 test dataset. The best results are highlighted and the second best results are underlined.}
\label{table3}
\resizebox{0.48\textwidth}{!}{
\begin{tabular}{@{}cllll@{}}
\toprule
Methods &  ACC (\%)     & AUC (\%)     & F1 (\%) & CPM (\%)\\
\midrule
MeanMIL & ${83.65}_{\pm5.29}$ & ${89.93}_{\pm1.37}$ & ${61.87}_{\pm7.34}$ & ${38.09_{\pm30.11}}$ \\
MaxMIL  & ${89.63}_{\pm0.62}$ & $\mathbf{{95.49}_{\pm0.30}}$ & ${47.78}_{\pm4.89}$ & \underline{$87.53_{\pm2.27}$} \\
ABMIL & ${38.77}_{\pm1.67}$ & ${61.22}_{\pm6.08}$ & ${31.17}_{\pm0.20}$ & $47.04_{\pm10.32}$ \\
CLAM & ${68.03}_{\pm18.69}$ & ${72.73}_{\pm15.65}$ & ${44.01}_{\pm10.14}$ & $50.19_{\pm16.87}$ \\
DSMIL & ${87.64}_{\pm0.49}$ & ${93.77}_{\pm0.43}$ & ${30.84}_{\pm4.73}$ & $71.85_{\pm12.85}$ \\
TransMIL & ${40.47}_{\pm3.58}$ & ${68.75}_{\pm2.17}$ & ${32.57}_{\pm1.16}$ & $66.67_{\pm1.05}$\\
DTFD & ${63.37}_{\pm21.41}$ & ${83.14}_{\pm7.17}$ & ${47.42}_{\pm18.10}$ & $75.56_{\pm3.43}$ \\
BCL(bag) & ${88.12}_{\pm3.66}$ & ${87.60}_{\pm4.06}$ & ${55.98}_{\pm7.57}$ & $83.46_{\pm1.23}$ \\
BCL(patch) & ${90.59}_{\pm4.23}$ & \underline{${94.27}_{\pm0.84}$} & \underline{${70.90}_{\pm6.69}$} & $-$\\
Ours(bag) & \underline{${91.48}_{\pm0.54}$} & ${90.14}_{\pm2.61}$ & ${60.80}_{\pm3.55}$ & 
$\mathbf{96.79_{\pm0.97}}$ \\
Ours(patch)  & $\mathbf{{93.88}_{\pm0.52}}$ & ${92.93}_{\pm2.87}$ & $\mathbf{{75.05}_{\pm3.06}}$ & $-$\\
\bottomrule
\end{tabular}
}
\end{table}

\begin{figure}[ht]
    \centering
    \includegraphics[width=\linewidth]{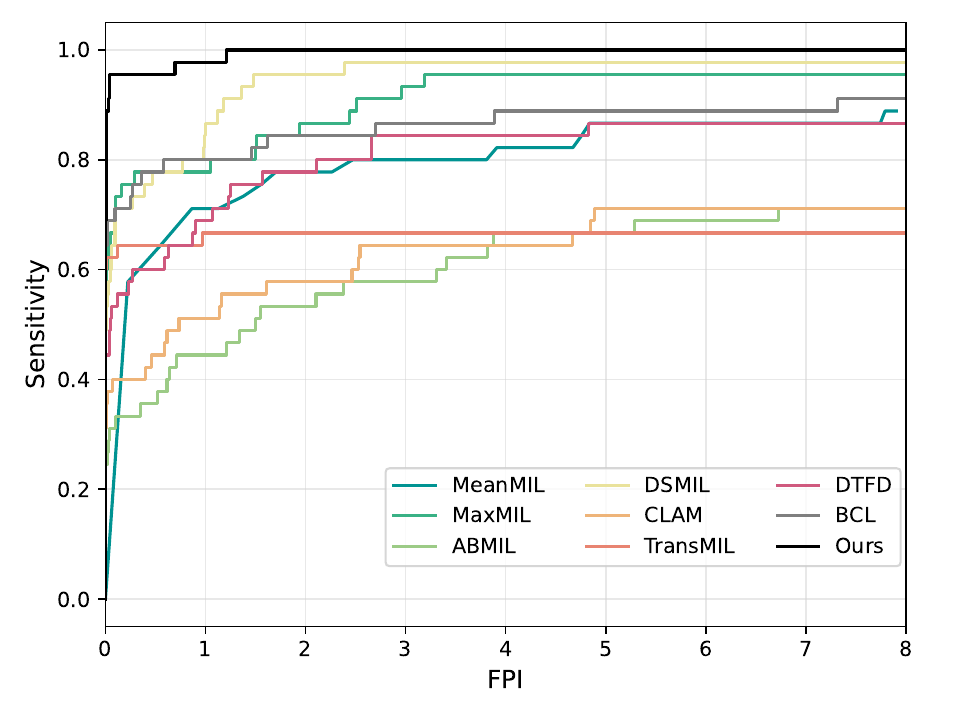}
    \caption{Free-response receiver operating characteristic curves with various MIL methods according to patch-level CAMELYON16 annotations.}
    \label{figure4}
\end{figure}
To further validate the WSI classification performance of our approach, we evaluated results across various MIL methods, as summarized in Tab. \ref{tab2}. Our analysis reveals that our approach consistently outperforms others on both datasets. Most of the other methods primarily focus on the improvement of the aggregator, yet they fail to achieve comparable performance to ours. This can be attributed to their reliance on features extracted by weights pre-trained on natural images, which is not optimal for WSI classification tasks. WSI-FT, an end-to-end training framework with partial patches, tends to exhibit poor generalization of features. Although BCL offers an excellent foundational framework, the presence of noise in the pseudo labels may potentially lead to model overfitting.
\begin{figure*}[ht]
    \centering
    \includegraphics[width=0.95\linewidth]{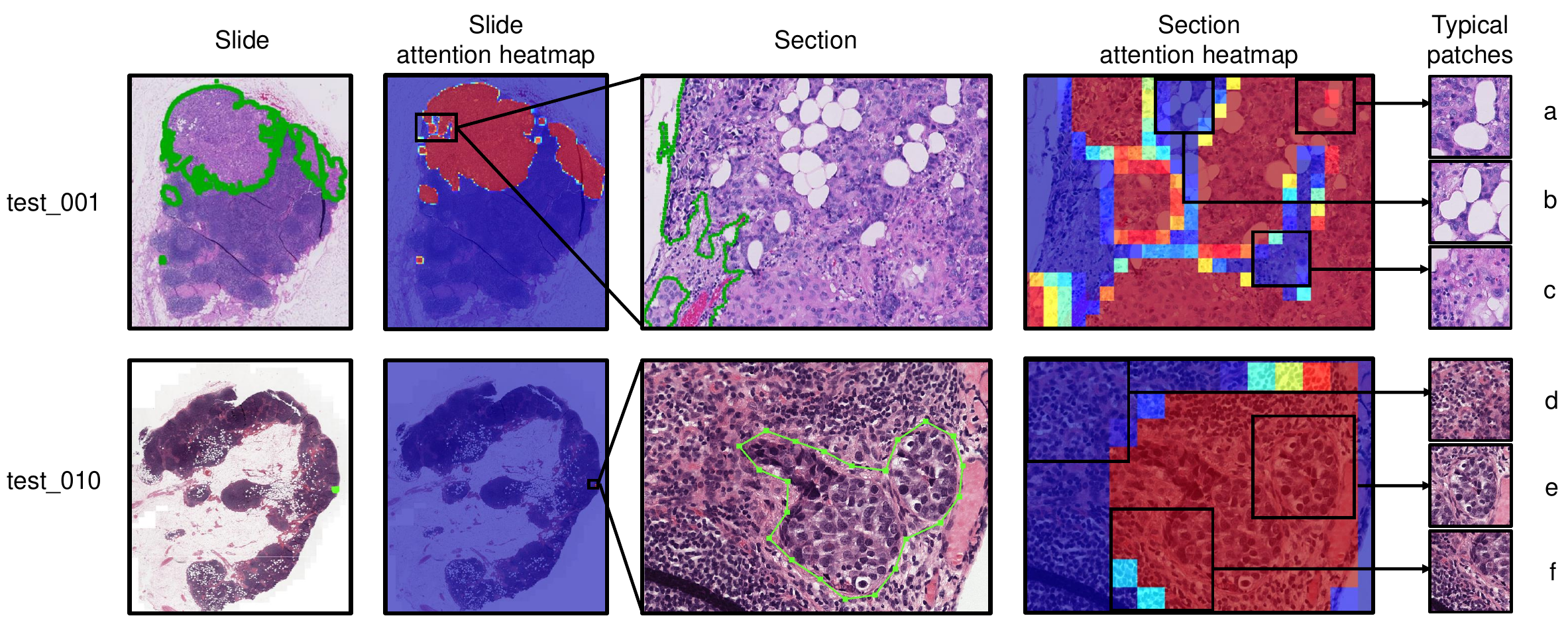}
    \caption{Heatmap of our method in the CAMELYON16 dataset. 'test$\_$001' and 'test$\_$010' refer to macro-metastasis and micro-metastasis WSIs. The green lines in the first column indicate the annotated regions of WSIs. The second column presents heatmaps generated by attention scores. The third and fourth columns display sections of the annotated area and the corresponding heatmaps. Typical patches (a)-(f) in the final column are selected for further analysis.}
    \label{figure5}
\end{figure*}

Both as iterative methods, we demonstrated the performance difference between BCL and our approach across different rounds of iteration on the validation and test sets of CAMELYON16 and BRACS datasets. Our observations in Fig. \ref{figure3} revealed that our method demonstrates a more rapid convergence compared to BCL on the CAMELYON16 dataset, accompanied by a notable improvement in effectiveness. On the BRACS dataset, although BCL's performance exhibits fluctuations, our method consistently converges swiftly to superior outcomes.

Given the detailed annotations for tumor regions provided by the CAMELYON16 dataset, we evaluated the patch-level classification ability of various MIL methods for comparison. All MIL methods were assessed using their bag-level classification classifiers, while both BCL and our method included an additional patch-level classifier. As presented in Tab. \ref{table3}, the performance results revealed that most methods exhibited high AUC scores alongside low F1 scores, which primarily stemmed from the inherent characteristics of the MIL principle. These methods tend to focus on a few instances within a bag, as a bag is labeled positive even if only one instance within it is positive. Consequently, this often leads to only a few patches being classified with very high confidence. In contrast, our method consistently achieved optimal results in both ACC and F1 scores. This can be attributed to our approach of selecting hard negative samples to mitigate overfitting. The performance of our bag-level classifier implicitly demonstrated that our method effectively addressed the tendency of MIL to excessively focus on a minority of patches. Furthermore, we provide free-response receiver operating characteristic curves using different MIL models, as depicted in Fig. \ref{figure4}. These curves illustrated that our method outperformed others across all FPIs, indicating the ability to detect tumor slides at a low false positive cost.

\subsection{Visualization and interpretation}
To provide an intuitive demonstration of the effectiveness of our method, we visualized representative results in Fig. \ref{figure5}. Specifically, we selected a macro-metastasis ('test$\_$001') and a micro-metastasis ('test$\_$010') WSIs from the CAMELYON16 dataset for visualization, generating classification heatmaps using attention scores derived from the MIL aggregator. For 'test$\_$001', our heatmap accurately identifies the macro-metastatic tumor region in the slide, along with some micro-metastatic tumor regions in the surrounding area (e.g., the lower-left corner). Upon closer analysis of regions with low attention in the heatmap, we observed that patches containing a large number of fat cells (e.g., patch ($b$)) were assigned lower attention. However, it is noteworthy that not all patches containing fat cells were dismissed as normal; for instance, in patches where only a small portion contains fat cells (e.g., patch ($a$)), our model still identifies them as tumors, indicating its ability to distinguish between fat cells and tumor cells. Additionally, regions with low attention also contained some lymph cells, which our model treats as hard negative samples in the dataset (e.g., patch ($c$)), leading to their lower attention assignment. For 'test$\_$010', our model successfully identifies the small tumor region, in accord with the annotation. Even in regions with micrometastatic tumor cells (e.g., patch ($f$)), our model demonstrates accurate identification. It's noted that since the patches are $256 \times 256$ in size, our heatmap slightly extends beyond the annotated tumor region.
\begin{table*}[ht]
\centering
\caption{Ablation study on potential negative sample mining, positive sample cleaning, and hard negative sample searching. The best results are highlighted. POS: positive sample; Mining: potential negative sample mining; Searching: hard negative sample searching; Cleaning: positive sample cleaning.}
\label{tab4}
\begin{tabular}{@{}ccccccccccc@{}}
\toprule
\multirow{2}{*}{Method} & \multirow{2}{*}{POS}  &  \multicolumn{3}{c}{Components}  &  \multicolumn{2}{c}{CAMELYON16} & \multicolumn{2}{c}{BRACS} &  \multicolumn{2}{c}{Average (\%)} \\ 
  &  & Mining & Searching & Cleaning & ACC (\%)     & AUC (\%)   & ACC (\%)     & AUC (\%)   & ACC    & AUC \\ 
\midrule
baseline & \checkmark &  &  & & ${76.35}_{\pm1.17}$ & ${74.51}_{\pm1.32}$ & ${57.85}_{\pm1.08}$ & ${80.06}_{\pm0.82}$ & 67.10 & 77.28\\
\midrule
\multirow{4}{*}{Ours}  & \checkmark &\checkmark &\checkmark & & ${80.24}_{\pm1.81}$ & ${79.15}_{\pm3.00}$ & ${58.62}_{\pm0.00}$ & ${80.61}_{\pm0.14}$ & 69.43 & 79.88\\
  & \checkmark &  & \checkmark & & ${76.75}_{\pm0.78}$ & ${80.60}_{\pm1.16}$ & ${63.22}_{\pm2.30}$ & ${81.95}_{\pm2.11}$ & 69.98 & 81.28\\
  & \checkmark &  & & \checkmark & ${77.13}_{\pm1.94}$ & ${74.99}_{\pm8.10}$ & ${59.20}_{\pm2.36}$ & ${78.84}_{\pm1.70}$ & 68.16 & 76.92\\
  & \checkmark & \checkmark & \checkmark & \checkmark & $\mathbf{{85.01}_{\pm1.32}}$ & $\mathbf{{90.35}_{\pm1.00}}$ & $\mathbf{{63.60}_{\pm2.36}}$ & $\mathbf{{82.16}_{\pm1.92}}$ & \textbf{74.31} & \textbf{79.88}\\

\bottomrule
\end{tabular}
\end{table*}

\begin{figure*}[ht]
    \centering
    \includegraphics[width=0.8\textwidth]{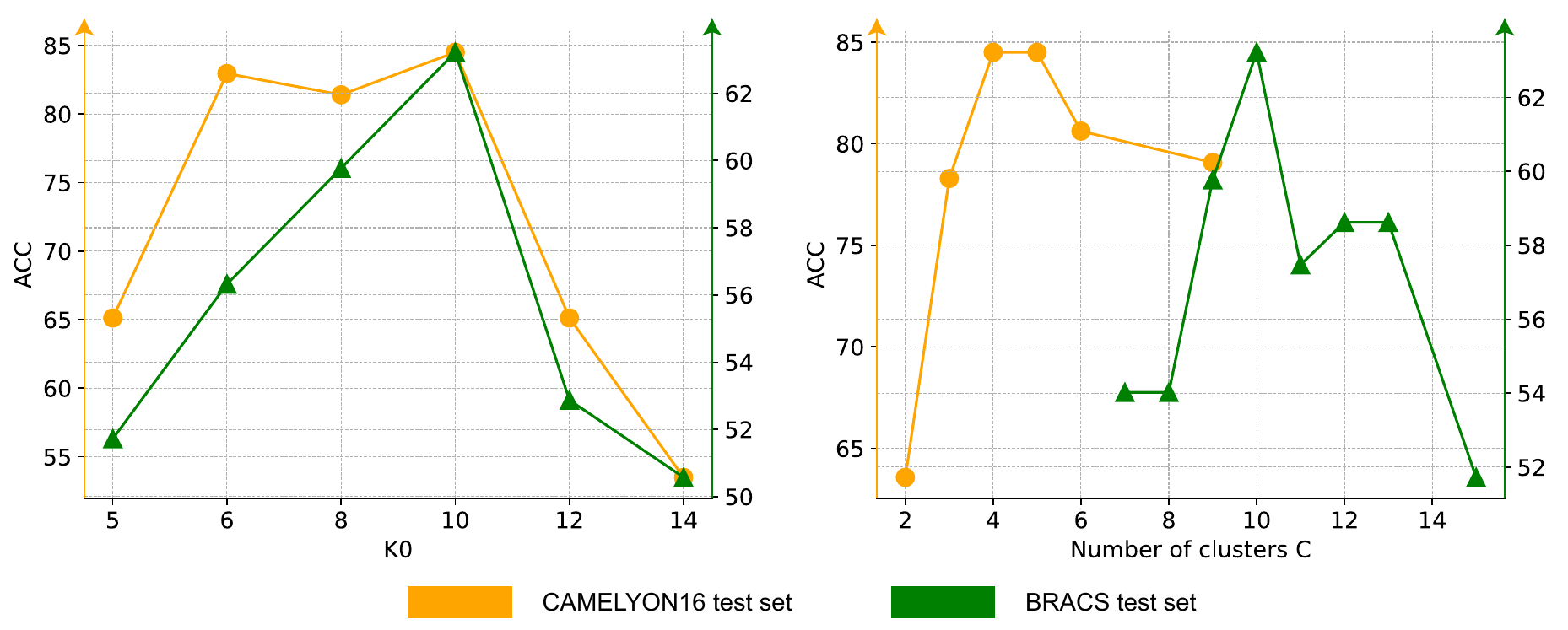}
    \caption{Results of our method in terms of ACC on the test set of CAMELYON16 and BRACS by varying the parameter $K_0$ and the number of clusters $C$.}
    \label{par}
\end{figure*}
\subsection{Ablation study}
To validate the effectiveness of key components within our method, we conducted ablation experiments on both the CAMELYON16 and BRACS datasets. As is depicted in Tab. \ref{tab4}, scrutinized three crucial components: potential negative sample mining, hard negative sample searching, and positive sample cleaning. For the potential negative sample mining component, compared to using only positive samples, a noticeable improvement was observed on the CAMELYON16 dataset. This can be attributed to the dataset's simplicity, which facilitated more accurate category center selection, thus enhancing the mining module's functionality. Conversely, on the BRACS dataset, the initial suboptimal MIL model generates less accurate category centers, thus consequently fewer useful negative samples are found for fine-tuning. In scenarios without hard negative sample searching, relying solely on high-confidence patches as hard negative samples showed mediocre performance on the CAMELYON16 dataset. In contrast, on the BRACS dataset, significant improvements were observed, as some negative samples were misclassified, thereby providing opportunities for enhancement. For experiments focusing solely on positive sample cleaning, both datasets exhibited slight improvements, indicating the presence of noise in the original positive samples. Notably, our method, although divided into three parts, operates synergistically: the negative sample mining module enables the model to identify missing information from low-confidence patches and maximize the utilization of all patches; the hard negative sample searching module aids the model in learning more adversarial information; and the positive sample cleaning module ensures the purity of positive samples, preventing the model from overfitting to noisy samples.

\subsection{Hyper-parameter study}
In this study, two critical hyper-parameters, namely the initial number of patches $K_0$ and the number of clusters $C$ are used. We conducted extensive experiments to set the best values for these two hyper-parameters. Limit our comparison to the outcomes following the initial iteration, as a selection of hyper-parameters. As depicted in Fig. \ref{par}, the curves all exhibit a trend of rising first and then declining. Therefore, we choose the parameters corresponding to the peak values as hyper-parameters. Specifically, we set $K_0$ to 10 on both datasets. As for the number of clusters $C$, since it largely depends on the dataset itself, we set it to $5$ for the CAMELYON16 dataset and $10$ for the BRACS dataset respectively. 

\section{Discussion and conclusion}
While multiple instance learning has been widely applied in WSI classification, most existing methods have hardly focused on updating the feature encoder. However, despite efforts to enhance the feature encoder, MIL models have often struggled to effectively aggregate suboptimal embeddings for final predictions. 

To address this limitation, we introduced HC-FT which utilizes the heuristic clustering strategy for feature fine-tuning. This heuristic clustering strategy enables the identification of feature centers for each category. Leveraging these centers, we developed two modules: the potential negative sample mining module enables the extraction of valuable information from areas not prominently focused on by the model; the label refinement module aids in the identification of hard negative samples and the purification of positive samples. We further fine-tune the feature encoder using selected patches with pseudo labels. Our results demonstrate that features generated using our method consistently exhibit superior performance, even when applied to the simplest ABMIL method.

However, our method has several shortcomings. Firstly, the result of the initial round is crucial, a superior initial outcome enables an effective identify category center. Conversely, if the initial result is poor, clustering centers may be inaccurate, rendering poor performance of subsequent modules. Moreover, due to the task-oriented nature of our approach, directly transferring our results to similar datasets remains uncertain. Lastly, we solely utilized classification tasks to update the feature encoder, other pretext tasks could potentially serve to update the feature encoder, meriting further exploration.

\section*{Acknowledgments}
This study was supported by Shenzhen Engineering Research Centre (XMHT20230115004), Science and Technology Research Program of Shenzhen City (KCXFZ20201221173207022), and Jilin Fuyuan Guan Food Group Co., Ltd.

\bibliographystyle{model2-names.bst}\biboptions{authoryear}
\bibliography{reference}

\end{document}